\documentclass{article}

\usepackage{microtype}
\usepackage{graphicx}
\usepackage{subcaption}
\usepackage{booktabs} 

\usepackage{hyperref}

\usepackage[preprint]{icml2026}

\usepackage{amsmath}
\usepackage{amssymb}
\usepackage{mathtools}
\usepackage{amsthm}
\usepackage{multirow}
\usepackage[table]{xcolor}
\usepackage{mdframed}
\usepackage{xcolor}
\definecolor{verbgray}{gray}{0.95}  
\newcommand{\red}[1]{\textcolor{black}{#1}}
\usepackage[capitalize,noabbrev]{cleveref}

\theoremstyle{plain}

\theoremstyle{definition}

\theoremstyle{remark}

\usepackage[textsize=tiny]{todonotes}

\icmltitlerunning{PhyScene3D: Physically Consistent Interactive 3D Tabletop Scene Generation}

\begin{document}

\twocolumn[
  \icmltitle{PhyScene3D: Physically Consistent Interactive 3D Tabletop Scene Generation}



  \icmlsetsymbol{equal}{*}
  \icmlsetsymbol{corr}{$\dagger$}

  \begin{icmlauthorlist}
    \icmlauthor{Weixing Chen}{equal,sysu}
    \icmlauthor{Zhuoqian Feng}{equal,sysu}
    \icmlauthor{Yang Liu}{sysu,bigdata}
    \icmlauthor{Yexin Zhang}{sysu}
    \icmlauthor{Yifan Wen}{sysu}
    \icmlauthor{Yinghong Liao}{comp}
    \icmlauthor{Weichao Qiu}{comp}
    \icmlauthor{Guanbin Li}{sysu,bigdata}
    \icmlauthor{Liang Lin}{corr,sysu,bigdata,pcl}
  \end{icmlauthorlist}

  \icmlaffiliation{sysu}{Sun Yat-sen University, China}
  \icmlaffiliation{pcl}{Peng Cheng Laboratory}
  \icmlaffiliation{bigdata}{Guangdong Key Laboratory of Big Data Analysis and Processing}
  \icmlaffiliation{comp}{Huawei}
  
  \icmlcorrespondingauthor{Liang Lin}{linliang@ieee.org} 
  

  \icmlkeywords{scene generation, Test-Time Optimization, Physically Grounded Reasoning and Alignment}

  \vskip 0.3in
]



\printAffiliationsAndNotice{}  

\begin{abstract}
Generating physically consistent 3D tabletop scenes is a fundamental yet underexplored problem for interactive and generalist robotic learning. 
The challenge stems from dense object hierarchies and irregular affordances. Here, an interactive scene denotes a physically valid, collision-free environment directly loadable into physics simulators.
Existing methods, ranging from decoupled symbolic solvers to end-to-end regression models, often suffer from error propagation or overfitting to noisy supervision containing widespread physical violations. 
To address these limitations, we introduce \textbf{PhyScene3D}, a framework that reformulates generation as a \textit{Human-Mimetic Constructive Process}. 
The proposed \textbf{Cognitive Topological Reasoning Chain (CTRC)} factorizes scene synthesis into a sequential, anchor-conditioned process. It employs a 3D AABB-based placement scheme that imposes a strong structural inductive bias.
To address imperfect supervision and physical infeasibility, we introduce \textbf{Physics-Aware Denoising Alignment (PADA)}. It integrates a differentiable Signed Distance Field (SDF) with Test-Time Optimization (TTO) to project generated scenes onto a physics-feasible manifold while preserving semantic intent.
Experiments demonstrate that PhyScene3D outperforms state-of-the-art approaches in both semantic accuracy and physical validity, achieving a 40\% reduction in scene-wise collision rate relative to the human-annotated training data.
\end{abstract}
\section{Introduction}

\begin{figure*}
    \centering
\includegraphics[width=0.93\linewidth]{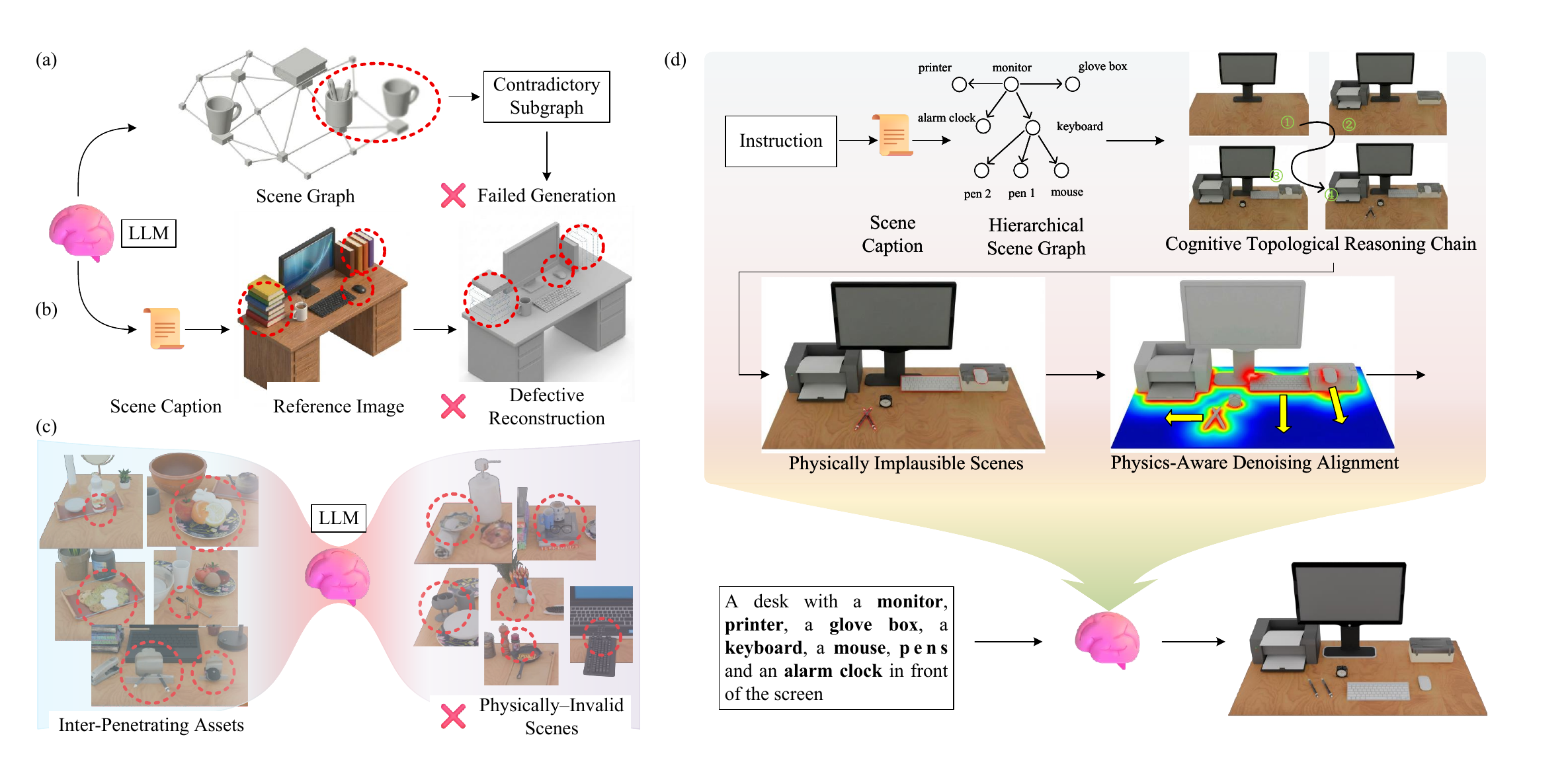}
    \caption{The current paradigms for desktop scene generation: (a) agent-based solvers, (b) image-based approaches, (c) end-to-end models, and (d) our proposed PhyScene3D.}
    \label{fig:intro}
\end{figure*}

The generalization capability of robotic agents in manipulation tasks hinges on their ability to perceive complex environments and execute actions grounded in human instructions~\cite{liu2025aligning}. To cultivate such generalist agents, researchers require simulation environments that reflect diverse physical causalities and semantic richness~\cite{torne2024reconcilingrealitysimulationrealtosimtoreal}, i.e., the interactive scenes\footnote{kinematically and physically valid, free of interpenetration, directly importable into physics engines \cite{xiang2020sapien,makoviychuk2021isaac}}. However, traditional scene construction methods, relying on manual heuristics~\cite{raistrick2024infinigen} or random arrangements~\cite{jia2024cluttergen}, fail to provide the scalability required for robust policy learning. Consequently, automated scene generation driven by Large Language Models (LLMs) has emerged as a promising paradigm, leveraging the extensive world knowledge embedded in pre-trained models~\cite{yang2024holodeck}.

While progress has been made in generating sparse indoor scenes, tabletop manipulation remains among the most challenging subsets of indoor environment synthesis, characterized by dense object hierarchies and irregular affordances~\cite{haomesatask, wang2025tabletopgen}. Unlike room-level furniture arrangement on a 2D ground plane, desktop organization demands rigorous 3D spatial topology, e.g., a pen must rest inside a holder that sits on a book. These complex physical dependencies expose fundamental limitations in existing paradigms.

Agent-based solvers~\cite{yang2024holodeck,ccelen2024design} attempt to address this by decoupling semantic planning (LLM) from geometric grounding (Solver), as shown in Figure \ref{fig:intro}(a). However, this creates a ``symbolic bottleneck": lacking fine-grained spatial awareness, the LLM often generates physically infeasible graphs (e.g., floating stacks) that downstream solvers cannot resolve without breaking the semantic intent~\cite{sun2025layoutvlm}. Conversely, image-mediated pipelines suffer from high latency and error accumulation inherent in their multi-stage ``generate-parse-retrieve" workflow (Figure \ref{fig:intro}(b)).

Direct end-to-end regression models provide an alternative but face a fundamental ceiling: blind mimicry. Since real-world datasets are rife with noise, ranging from physical collisions to unstable placements, naive supervised learning forces models to copy these patterns rather than learning the underlying physical laws (Figure \ref{fig:intro}(c)). Consequently, they fail to generate the high-fidelity, collision-free environments required for reliable robotic simulation.

To overcome these limitations, we propose \textbf{PhyScene3D}, a framework that rethinks generation not as distribution matching, but as a ``Human-Mimetic Constructive Process". 
We argue that internalizing the explicit planning pipeline into the VLM's implicit reasoning activates its geometric priors. 
Crucially, we treat the noisy training data not as a ground truth to be replicated, but as an imperfect reference to be ``denoised". By integrating a differentiable physical signal, the model is enabled to transcend its supervision and learn layouts that are physically more plausible than the human demonstrations on which it was trained.

We realize this goal through two core mechanisms (Figure \ref{fig:intro}(d)). First, the \textbf{Cognitive Topological Reasoning Chain (CTRC)} acts as a structural sequencer. It reformulates flat generation into an anchor-based hierarchical planning pipeline. By mapping abstract Scene Graphs to the 3D \textbf{AABB-based placement scheme}, CTRC imposes a strong inductive bias that enforces geometric dependencies (e.g., placing containers before their contents), effectively eliminating causal hallucinations. 
Second, to guide the planner through noisy data, we introduce \textbf{Physics-Aware Denoising Alignment (PADA)}. Unlike standard reinforcement learning, which often leads to ``semantic drift" (sacrificing logic for collision avoidance), PADA employs a \textbf{Hybrid Optimization} strategy. It utilizes a differentiable SDF engine to perform Test-Time Optimization (TTO) on model hypotheses, generating physically projected semantic anchors that guide the policy. This approach enables PhyScene3D to rigorously optimize physical plausibility while faithfully adhering to the user’s semantic intent.

Our contributions are threefold:
\begin{itemize}
\setlength{\itemsep}{0pt}
\setlength{\topsep}{0pt}
\setlength{\parsep}{0pt}

\item We propose PhyScene3D, the first framework to inject a differentiable 3D physics engine into the training loop of a vision-language model. This enables generative models to surpass the physical quality of their training data through self-supervised denoising.


\item We introduce Cognitive Topological Reasoning Chain (CTRC), a hierarchical sequencer that imposes a strong structural inductive bias by linearizing complex scene graphs into anchor-relative assembly sequences.


\item We develop Physics-Aware Denoising Alignment (PADA), a hybrid RL strategy that resolves the reward-hacking and semantic-drift trade-off by distilling differentiable SDF-based physical priors into the VLM through TTO-refined semantic anchors.

\end{itemize}
\section{Related Work}

\subsection{Interactive 3D scene Generation}

\begin{figure*}[h]
    \centering
    \includegraphics[width=1\linewidth]{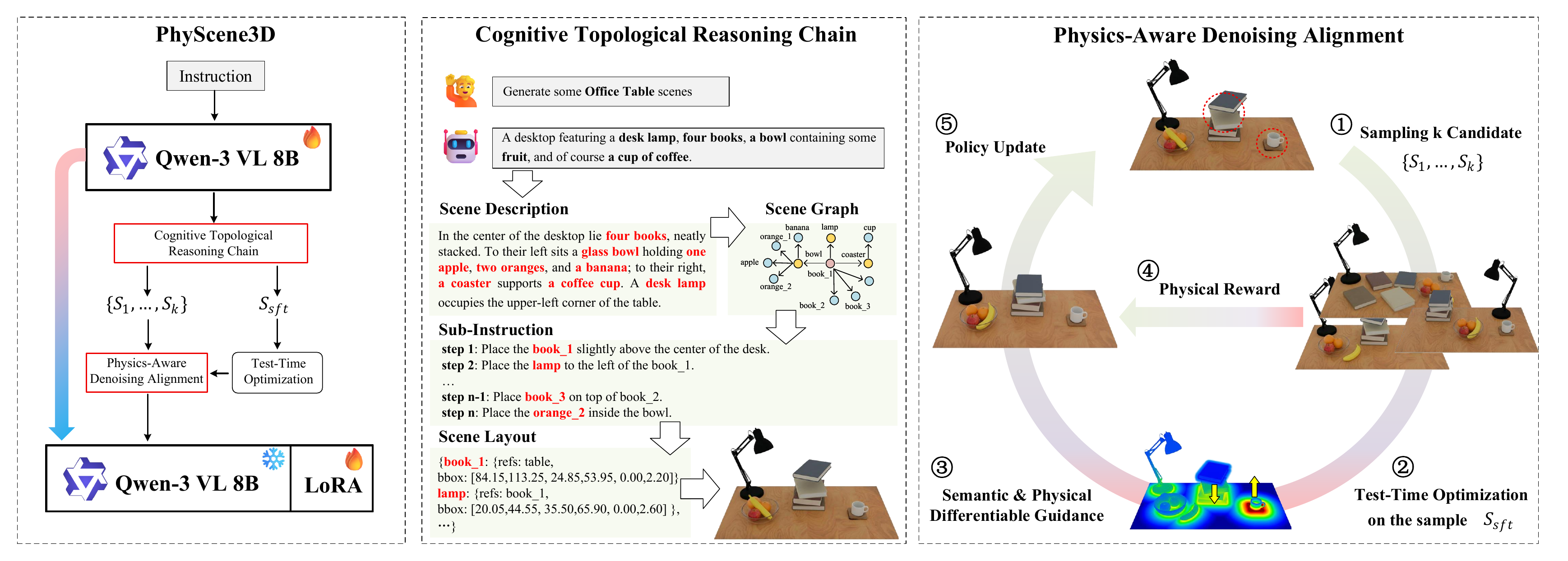}
    \caption{The PhyScene3D framework illustrates our two-stage training paradigm: (a) first, supervised fine-tuning equips the model with the Cognitive Topological Reasoning Chain mindset; (b) then, Physics-Aware Denoising Alignment further mitigates physical noise in the data to achieve superior semantic consistency and physical plausibility.}
    \label{fig:method}
\end{figure*}

Automated scene synthesis can be categorized into multi-stage agentic pipelines and end-to-end regression approaches. Text-centric agents, such as Holodeck~\cite{yang2024holodeck} and I-Design~\cite{ccelen2024design}, translate prompts into symbolic constraints. Yet, they suffer from a ``symbolic bottleneck", where the lack of spatial grounding of LLM causes geometrically infeasible hallucinations (e.g., intersecting volumes) that deterministic solvers cannot resolve~\cite{sun2025layoutvlm}. Image-mediated approaches~\cite{ling2025scenethesis, wang2025tabletopgen} attempt to bridge this by lifting generated or real-world 2D images~\cite{yu2025metascenes, El_Amine_Boudjoghra_2025_ICCV} to 3D. However, this ``generate-parse-retrieve" workflow introduces high latency and irreversible error propagation. Conversely, end-to-end methods like MesaTask~\cite{haomesatask} model joint distributions directly from data. While efficient, this paradigm is bounded by supervision quality: naive training encourages models to replicate dataset artifacts, including physical collisions. Unlike PhyScene3D, these methods lack intrinsic mechanisms to transcend imperfect demonstrations via physical self-correction.

\subsection{Physically Grounded Reasoning and Alignment}

While Vision-Language Models (VLMs) demonstrate spatial awareness in perceptual tasks~\cite{bai2025qwen3vltechnicalreport, hurst2024gpt}, leveraging these priors for constructive generation remains challenging. Current embodied AI research predominantly utilizes VLMs for semantic planning~\cite{driess2023palm}, visual navigation~\cite{huang2023visual}, or manipulation policies~\cite{huang2023voxposer}. Although recent efforts~\cite{chen2024spatialvlm, Jiang_2025_ICCV, karamcheti2024prismatic} explicitly tune models for spatial reasoning, they typically focus on analyzing existing scenes via visual input~\cite{wang2024robogen}. Deploying VLM geometric priors as an implicit planner without runtime visual cues remains underexplored. A critical barrier is aligning semantic reasoning with physical laws. Prior methods rely on non-differentiable rejection sampling via rigid simulators~\cite{todorov2012mujoco, xiang2020sapien} or heuristic filters~\cite{Paschalidou2021NEURIPS, tang2024diffuscene}, which sever gradient flow and force expensive trial-and-error. We bridge this gap via Physics-Aware Denoising Alignment (PADA). By integrating a Differentiable Signed Distance Field (SDF) into reinforcement learning, we provide dense, gradient-based feedback, enabling PhyScene3D to actively ``denoise" its policy against data imperfections.

\section{Methodology}
\label{sec:method}

We present \textbf{PhyScene3D}, a unified framework designed to bridge the gap between abstract user instructions and physically plausible, hierarchically dense 3D desktop environments. As illustrated in Figure~\ref{fig:method}, rather than treating scene generation as a static regression task, we reformulate it as a \textit{Human-Mimetic Constructive Process}.

The framework consists of two organically integrated modules:
1) \textbf{Cognitive Topological Reasoning Chain (CTRC)}: A sequencer that imposes a strong inductive bias by linearizing scene generation into an anchor-based, hierarchically dependent planning stream,
2) \textbf{Physics-Aware Denoising Alignment (PADA)}: A grounding mechanism that utilizes a differentiable physics engine to project semantic plans onto physically valid manifolds, enabling robust policy optimization without explicit semantic reward modeling.

\subsection{Problem Formulation}
Given a natural language instruction $\mathcal{I}$, our objective is to generate a desktop scene layout $\mathcal{S} = \{e_i\}_{i=1}^N$, where $N$ is the number of objects. We define each entity $e_i$ as a tuple $e_i = (c_i, \mathbf{p}_i, \mathbf{s}_i, \theta_i)$, where $c_i$ denotes the semantic category, $\mathbf{p}_i \in \mathbb{R}^3$ is the absolute center position, $\mathbf{s}_i \in \mathbb{R}^3$ represents spatial dimensions, and $\theta_i \in \mathbb{R}$ is the yaw rotation.

To represent the complex structural dependencies of the desktop scene layout and facilitate hierarchical reasoning detailed in Sec.~\ref{sec:ctrc}, we re-parameterize the explicit position $\mathbf{p}_i$ and dimensions $\mathbf{s}_i$ into a 3D Axis-Aligned Bounding Box (AABB) representation defined by coordinate extrema: $\mathbf{b}_i = [x_{\min}, x_{\max}, \dots, z_{\max}] \in \mathbb{R}^6$. Unlike the 6D pose and the object size commonly used in previous work~\cite{haomesatask}, this object representation unifies position and scale, allowing the model to explicitly predict spatial occupancy boundaries and measure the relative placement relations between the objects. Thus, the task is to model the conditional distribution $P(\mathcal{S} | \mathcal{I})$, mapping instructions to a sequence of semantic-geometric attributes.

\subsection{Cognitive Topological Reasoning Chain (CTRC)}
\label{sec:ctrc}

Naive generative models often treat objects as independent entities, resulting in ``floating" artifacts or geometric hallucinations. CTRC addresses this issue by emulating the human cognitive process of \textit{hierarchical decomposition}. It seamlessly integrates a topological graph structure with the 3D AABB-based placement scheme, culminating in a sequential planning strategy.

\subsubsection{Hierarchical Scene Graph (HSG)}
The foundation of our reasoning chain is the HSG, a directed acyclic graph $G=(V, E)$ that abstracts the scene into geometric dependencies rather than mere semantic labels. We extract edges $E$ based on geometric heuristics ($\text{In} \succ \text{On} \succ \text{Near}$):
\begin{itemize}

    \item \textbf{Containment ($\mathtt{in}$):} Defined by volume ratio ($V_B/V_A \geq 1.5$) and high footprint overlap ($IoU_{xy} \geq 0.9$), where A is content, and B is container.
    \item \textbf{Vertical Support ($\mathtt{on}$):} Defined by vertical adjacency ($z_{min}^A \approx z_{max}^B$) and stability.
    \item \textbf{Horizontal Proximity ($\mathtt{near}$):} Defined by the dominant separation axis relative to the table center.
\end{itemize}
This graph serves as the ``cognitive map", establishing that a \textit{pen} is not merely ``at $(x,y)$" but is ``inside the holder".

\subsubsection{3D AABB-based Placement Scheme}
Building upon the HSG, we introduce the 3D AABB-based placement scheme that enforces the graph's structural constraints. We define the pose of a child object $e_{ch}$ strictly in the local frame of its parent $e_{pa}$.
Let $\mathbf{b}^{abs}$ denote absolute coordinates. We compute relative coordinates $\mathbf{b}^{rel}$ as offsets. Crucially, we decouple the vertical dimension to encode physical affordance:
\begin{equation}
    \label{eq:z_rel}
    z^{rel}_{\{min, max\}} = \begin{cases} 
    z^{abs}_{\{min, max\}}(e_{ch}) - z^{abs}_{min}(e_{pa}) & \text{if } r = \mathtt{in} \\
    z^{abs}_{\{min, max\}}(e_{ch}) - z^{abs}_{max}(e_{pa}) & \text{if } r = \mathtt{on}
    \end{cases}
\end{equation}
This formulation creates \textit{geometric invariants}: even if the parent moves, the child's relative parameterization remains constant. This drastically reduces the search space, rendering 'inside' and 'on' as mathematically distinct relative placements. AABB here serves as a macro-topology inductive bias for hierarchical planning, while true physical fidelity is enforced in the subsequent PADA stage via gradients on high-precision mesh SDFs. For extremely irregular shapes such as baskets with handles, AABB may conservatively overestimate occupied volume; the SDF engine corrects this at fine-grained contact points.

\subsubsection{Anchor-Based Sequential Planning}
The core innovation of CTRC is utilizing the HSG and relative AABB to linearize generation into an \textbf{Anchor-Based Sequential Plan}. Drawing on cognitive science findings that humans organize spatial tasks hierarchically~\cite{tversky1981distortions}, we enforce a specific generation order:
\begin{equation}
    P(\mathcal{S}|\mathcal{I}) = \prod_{t=1}^N P(e_t | e_{<t}, \mathcal{I})
\end{equation}
The sequence $e_{1 \dots N}$ is deterministically ordered by our \textit{Anchor-Expansion Strategy}:
1.  \textbf{Anchor Identification:} The process begins with the ``Base Anchor", the object on the table closest to the center.
2.  \textbf{Horizontal Expansion:} We iteratively generate objects on the table surface, referencing existing neighbors (e.g., \textit{``Place the lamp left of the laptop''}).
3.  \textbf{Recursive Stacking:} Critically, once a container or support is generated, its children are immediately planned in a bottom-up sequence. 

This strategy provides a strong inductive bias that respects physical causality. During training, the generation order is deterministically derived from the scene graph via bottom-up geometric heuristics, ensuring supports precede their contents. In inference, the model autoregressively predicts this order from SFT priors without any external rule engine. Even if an incorrect order is predicted, the relative-coordinate parameterization immediately exposes the error through floating or penetration, allowing the SDF engine in PADA to penalize such violations during RL and force the policy to recover physical causality.

\subsection{Physics-Aware Denoising Alignment (PADA)}
\label{sec:pada}

While CTRC provides a structurally coherent plan, the raw predictions of VLMs are inherently bounded by the ``physical noise'' (e.g., collisions, levitation) present in the training data. PADA addresses this by grounding the VLM via a differentiable physics engine. 

However, a critical limitation exists: optimizing solely for physical validity (e.g., zero collision) often induces \textit{semantic drift}, where the model satisfies physical constraints by destroying the intended arrangement (e.g., scattering objects far apart). To resolve this, we propose a hybrid optimization strategy where the \textbf{3D AABB-based placement scheme} (Sec.~\ref{sec:ctrc}) acts as a rigid constraint, ensuring that physical correction does not compromise semantic fidelity.
\subsubsection{Differentiable SDF Physics Engine}
The cornerstone of our approach is a GPU-resident physics engine that preprocesses all assets into vectorized Signed Distance Fields (SDFs), denoted as $\phi$. This engine serves a dual purpose: enabling fast reward computation for reinforcement learning (RL) and providing differentiable gradients for optimization. Given a layout, for each object $A$, we derive its world-space rotation $\mathbf{R}_A$ and translation $\mathbf{t}_A$. The collision energy against a target SDF $\phi_B$ is computed via differentiable point sampling:
\begin{equation}
\label{eq:sdf}
    \mathcal{L}_{sdf}(A, B) = \sum_{\mathbf{p} \in P_A} \text{ReLU}\left( -\phi_B\left( \mathbf{R}_B^\top (\mathbf{R}_A \mathbf{p} + \mathbf{t}_A - \mathbf{t}_B) \right) \right)
\end{equation}
where $P_A$ denotes surface points sampled in $A$'s canonical frame. The pose parameters are thus optimized by the gradients from the physics engine, physically ``pushing'' the intersecting objects apart.

\subsubsection{Hierarchy-Constrained Test-Time Optimization (TTO)}
To utilize this engine without breaking semantics, we introduce a Hierarchy-Constrained TTO. Standard TTO optimizes absolute coordinates, which is prone to drift. Instead, we leverage the 3D AABB-based placement scheme established in CTRC to enforce structural rigidity.

We treat the VLM's output as an initialization and optimize the geometric parameters $\xi$. The objective function strictly penalizes deviations in the \textit{relative} frame rather than the global frame:
\begin{equation}
    \min_{\xi} \left( \mathcal{L}_{sdf} + \lambda_{rel} \mathcal{L}_{rel}(\mathcal{G}) + \lambda_{reg} ||\xi - \xi_{init}||^2 \right)
\end{equation}
Here, $\mathcal{L}_{rel}(\mathcal{G})$ imposes constraints based on the Scene Graph edge types:
\begin{itemize}
    \item For $\mathtt{in}$ edges, we freeze the relative position, treating the child and parent as a rigid body.
    \item For $\mathtt{on}$ edges, we enforce the $z^{rel}$ alignment constraint defined in Eq. \ref{eq:z_rel}, ensuring the object remains on the surface even if the surface moves to resolve collisions.
\end{itemize}
This formulation ensures that optimization resolves collisions by adjusting the assembly as a whole or making micro-adjustments within valid affordance zones, thereby preserving the semantic intent.

\subsubsection{Physically-Projected Semantic Anchoring}
Finally, we address the challenge of training the VLM to satisfy physical constraints without a dense semantic reward model. We propose a \textbf{Physically-Projected Semantic Anchor} strategy that generates high-quality pseudo-supervision from the SFT model itself.

\textbf{Rationale of Semantic Anchoring.}
The SFT model (trained on instruction data) captures the correct semantic distribution $\mathcal{M}_{sem}$ but suffers from physical noise. Conversely, pure RL exploration finds the physical manifold $\mathcal{M}_{phys}$ but lacks semantic direction.
We bridge this gap by leveraging the refinement capability of the TTO engine into the VLM's generative policy. Concretely, we generate a pseudo-label from the SFT model and refine it via TTO, creating a physically-projected semantic anchor that guides RL exploration without requiring the model to rediscover semantic structure from scratch:
\begin{enumerate}
    \item \textbf{Inference:} Generate a layout hypothesis $\mathcal{S}_{sft}$ using the SFT model given the training instruction.
    \item \textbf{Rectification:} Apply Hierarchy-Constrained TTO to project $\mathcal{S}_{sft}$ onto the physical manifold, yielding $\mathcal{S}^*_{anchor} = \text{TTO}(\mathcal{S}_{sft})$.
\end{enumerate}
Because TTO is constrained by the 3D AABB-based placement scheme, $\mathcal{S}^*_{anchor}$ retains the semantic fidelity of SFT while being collision-free.

\textbf{Dual-Objective RL.} 
We integrate this anchor into Group Relative Policy Optimization (GRPO) as a dense supervision term:
\begin{align}
    \mathcal{J}(\theta) = \mathbb{E}_{\mathcal{L} \sim \pi_\theta} & \underbrace{\left[ \frac{1}{K} \sum_{k=1}^K A_k \frac{\pi_\theta(\mathcal{S}_k)}{\pi_{old}(\mathcal{S}_k)} \right]}_{\text{RL: Exploration}} \nonumber \\
    & + \underbrace{\alpha \cdot \mathcal{L}_{CE}(\pi_\theta, \mathcal{S}^*_{anchor})}_{\text{Anchor: Semantic Stability}}
\end{align}

The RL term ($A_k$ derived from collision scores) drives the model to explore physical validity. The Anchor term provides a ``correction vector" pointing towards $\mathcal{L}^*_{anchor}$. This significantly accelerates convergence because the model does not need to rediscover semantic structures (e.g., ``cup holds pen") through random exploration; it only needs to learn the micro-adjustments required to satisfy physics. This effectively distills the analytical power of the TTO engine back into the VLM's priors.

\section{Experiments}
\label{sec:experiments}

\begin{table*}[]
    \centering
    \setlength{\tabcolsep}{5pt}
        \caption{The performance of Layout generation. \colorbox{gray!10}{Reference} denotes the effectiveness of the benchmark constructed from samples drawn from MesaTask-10k, serving as a baseline to measure the inherent quality of the data. We present the performance of three approaches, including \colorbox{blue!10}{Zero-shot by closed-source LLM}, \colorbox{orange!10}{Agent-Based Solver}, and \colorbox{green!10}{End-to-End Model}. Bold indicates the best performance, and underline indicates the second-best.}
    \begin{tabular}{lccccccccccc}
    \toprule
    \multirow{2}{*}{Model}  & \multicolumn{3}{c}{Quality Pass Rate} & \multicolumn{6}{c}{GPT Score} & \multicolumn{2}{c}{Collision Ratio} \\
    \cmidrule(lr){2-4}\cmidrule(lr){5-10}\cmidrule(lr){11-12}
    & $\tau=7$ & $\tau=8$ & $\tau=9$ & CwT & OSR & PPI & LCR & OV & Avg. & Scene-wise$\downarrow$ & Asset-wise$\downarrow$\\
    \midrule
    \rowcolor{gray!10}
    Reference & 17.1\% & 15.1\% & 11.5\% & \underline{7.83} & \underline{9.25} & 9.52 & \underline{8.25} & \textbf{9.18} & \underline{8.87} & 81.5\% & 8.19\%\\
    \midrule
    \rowcolor{blue!10}
    GPT-4o-mini & 17.9\% & 13.4\% & 8.8\% & 6.18 & 8.45 & 9.15 & 6.82 & 7.34 & 7.59 & 80.0\% & 13.61\% \\
    \rowcolor{blue!10}
    GPT-4o  & \underline{27.6}\% & \underline{22.1}\% & 12.7\% & 6.95 & 8.79 & \underline{9.54} & 7.48 & 8.18 & 8.19 & 68.9\%  &  7.87\%\\
    \midrule
    \rowcolor{orange!10}
    Holodeck-table & 2.7\% & 0.6\% & 0\% & 2.35 & 6.21 & 7.38 & 3.61 & 3.43 & 4.60 & \textbf{2.7}\% & \textbf{0.47}\%\\
    \rowcolor{orange!10}
    I-Design-table & 19.0\% & 9.7\% & 4.5\% & 4.39 & 8.20 & 9.20 & 5.34 & 5.55 & 6.53 & \underline{39.1}\% & 5.94\% \\
    \midrule
    \rowcolor{green!10}
    MesaTask & 21.1\% & 18.6\% & \underline{14.8}\% & 7.78 & 9.23 & 9.78 & 8.15 & 9.05 & 8.80 & 78.3\% & 8.19\%\\
    \rowcolor{green!10}
    Ours & \textbf{46.5}\% & \textbf{39.1}\% & \textbf{27.3}\% & \textbf{7.85} & \textbf{9.45} & \textbf{9.91} & \textbf{8.29} & \underline{9.15} & \textbf{8.93} & 41.6\% & \underline{3.86}\%\\
    \bottomrule
    \end{tabular}
    \label{tab:main_result}
\end{table*}


\subsection{Experimental Setup}

\textbf{Dataset.} 
We introduce \textbf{MesaTask-CTRC}, derived from the MesaTask-10k dataset~\cite{haomesatask} by augmenting raw layouts with hierarchical scene graphs and goal-oriented instructions (construction details in Appendix~\ref{app:dataset}). The dataset comprises 9,429 training samples and a challenging benchmark of 866 samples. The benchmark is stratified across 6 distinct tabletop scenarios and 5 difficulty levels based on object density and stacking complexity. The Reference data exhibits an 81.5\% SCR, reflecting genuine human-annotated interpenetration rather than AABB false positives.

\textbf{Baselines.} 
We compare PhyScene3D against three distinct paradigms:
(1) Zero-shot Generation with Closed-source LLMs: It employs state-of-the-art models, including \textbf{GPT-4o} and \textbf{GPT-4o-mini}, for zero-shot scene generation.
(2) Agent-Based Solver: Methods such as \textbf{Holodeck}~\cite{yang2024holodeck} and \textbf{I-Design}~\cite{ccelen2024design}, which leverage LLMs to generate symbolic constraints for external solvers.
(3) End-to-End Model: The supervised baseline from \textbf{MesaTask}~\cite{haomesatask}, which regresses object poses directly without hierarchical intermediates.

\textbf{Implementation Details.} 
We utilize \textbf{Qwen-3 VL 8B}~\cite{bai2025qwen3vltechnicalreport} as the backbone, trained on devices with 640 GB VRAM in total. Training proceeds in two stages:
\textit{i) Supervised Fine-Tuning (SFT):} Full-parameter optimization on the training set.
\textit{ii) Physically-Aware RL:} GRPO with LoRA adapters ($r=16$) to inject physical gradients. 
Detailed hyperparameters for learning rates, batch sizes, and TTO coefficients are provided in Appendix~\ref{app:implementation}.

\textbf{Metrics.} 
We evaluate performance on two dimensions:
\textbf{1) Physical Plausibility:} We report \textit{Scene-wise Collision Rate (SCR)}, the percentage of scenes with violations, and \textit{Asset-wise Collision Rate (ACR)}, the percentage of colliding object pairs. \textbf{2) Semantic Fidelity:} We employ \textit{GPT-Score} (1-10 scale) in 5 dimensions to assess layout logic and visual appearance, including \textit{Consistency with Task} (CwT), \textit{Object Size Reasonableness} (OSR), \textit{Placement Plausibility \& Intersections} (PPI), \textit{Layout Coherence \& Realism} (LCR), and \textit{Object Visibility} (OV). Finally, we propose an evaluation metric to measure data usability, namely \textbf{Quality Pass Rate (QPR)}. It measures the proportion of valid samples with a GPT Score greater than $\tau$ that are collision-free.

\subsection{Main Result}

\label{subsec:main_results}

\begin{table*}[h]
    \centering
    \setlength{\tabcolsep}{5pt}
        \caption{The performance of Layout generation on the OOD benchmark.}
    \begin{tabular}{lccccccccccc}
    \toprule
    \multirow{2}{*}{Model}  & \multicolumn{3}{c}{Quality Pass Rate} & \multicolumn{6}{c}{GPT Score} & \multicolumn{2}{c}{Collision Ratio} \\
    \cmidrule(lr){2-4}\cmidrule(lr){5-10}\cmidrule(lr){11-12}
    & $\tau=7$ & $\tau=8$ & $\tau=9$ & CwT & OSR & PPI & LCR & OV & Avg. & Scene-wise$\downarrow$ & Asset-wise$\downarrow$\\
    \midrule
    MesaTask & 1.01\% & 0.51\% & 0.51\% & 5.02 & 7.96 & 8.50 & 5.75 & 6.17 & 6.68 & 97\% & 19.01\%\\
    Qwen-3 VL 8B & 3.08\% & 1.54\% & 1.03\% & 6.51 & \textbf{8.88} & 9.44 & 7.01 & 7.77 & 7.92 & 96.4\% & 10.28\%\\
    + CTRC & 10.6\% & 8.04\% & 3.02\% & \textbf{6.74} & 8.86 & \textbf{9.50} & \textbf{7.34} & \textbf{7.96} & \textbf{8.08} & 88.4\% & 6.62\%\\
    + CTRC \& PADA & \textbf{29.1}\% & \textbf{15.1}\% & \textbf{7.03}\% & 6.19 & 8.63 & 9.46 & 6.83 & 7.56 & 7.56 & \textbf{63.8}\% & \textbf{2.92}\%\\
    \bottomrule
    \end{tabular}
    \label{tab:ood_result}
\end{table*}

We perform a multi-dimensional analysis to validate PhyScene3D's capability to generate physically plausible and semantically faithful environments.
Table~\ref{tab:main_result} presents a comprehensive evaluation of PhyScene3D against state-of-the-art baselines. The empirical evidence strongly supports our central hypothesis: by internalizing explicit structural planning and physical feedback, a generative model can surpass the quality of its noisy training supervision.

\subsubsection{Quantitative Analysis}
\label{subsubsec:quantitative}

A critical limitation in data-driven generation is blind mimicry, where models learn to reproduce dataset artifacts. As shown in Table~\ref{tab:main_result}, the Reference data, representing the upper bound of human annotation, suffers from an inherent SCR of 81.5\%. The End-to-End baseline (MesaTask) statistically mirrors this distribution. In contrast, PhyScene3D significantly reduces the ACR to \red{3.86}\%, a relative reduction of nearly 50\% compared to the ground truth. Crucially, this physical regularization does not come at the cost of semantic expressiveness. Our method achieves a QPR ($\tau=7$) of \red{46.5}\%, surpassing the Reference data itself (17.1\%) by a wide margin. Analyzing the GPT-Score dimensions reveals that this gain is driven by precision: PhyScene3D outperforms baselines in the PPI score ({9.91} vs. 9.54 for GPT-4o) and OSR score ({9.45} vs. 9.25 for Reference), validating that the PADA module effectively ensures physically realistic predictions.

\begin{figure}
    \centering
    \includegraphics[width=1\linewidth]{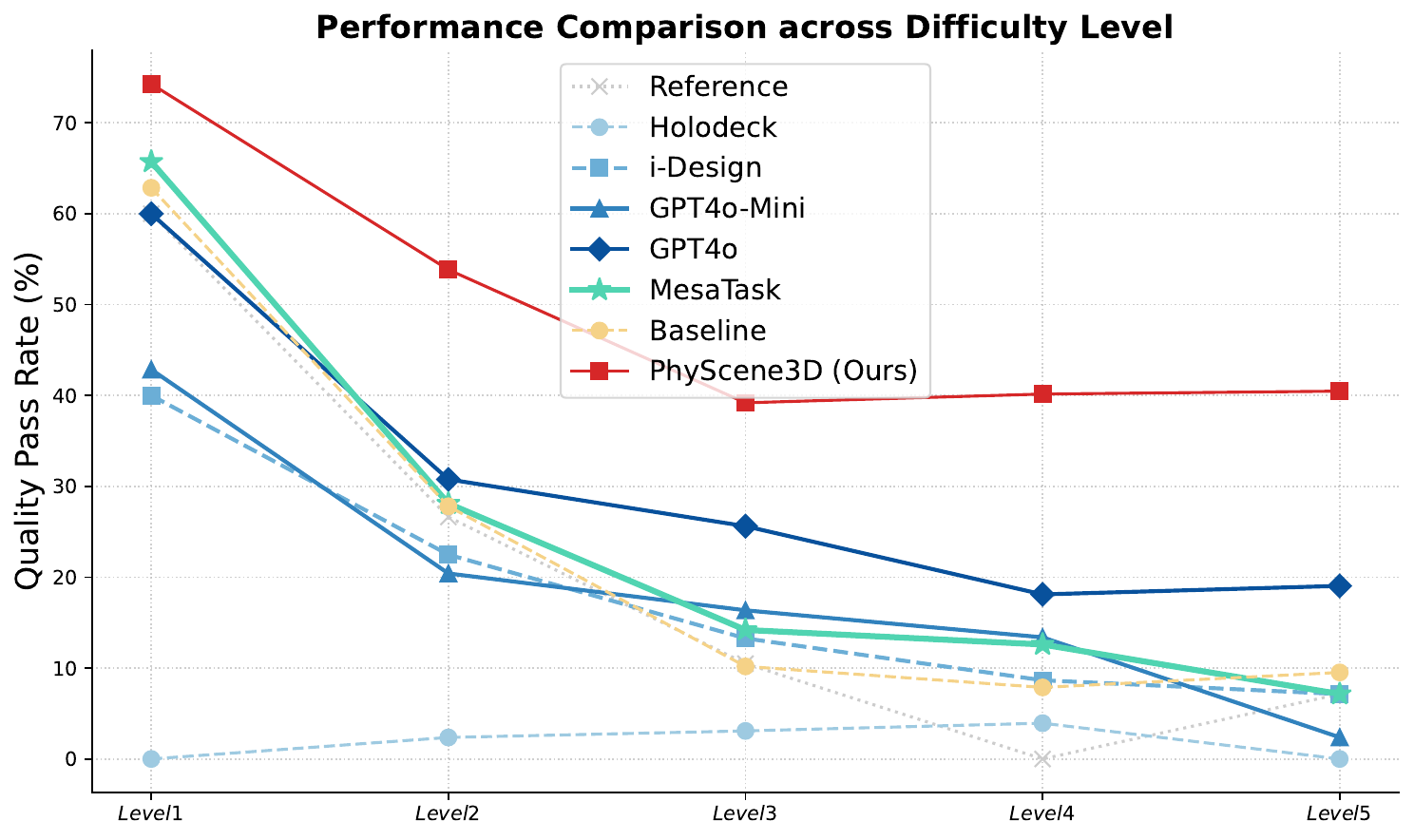}
    \caption{QPR ($\tau=7$) performance of each method across different difficulty levels.}
    \label{fig:difficulty_curve}
\end{figure}
\begin{table}
    \centering
    \caption{Comparison between the diffusion model-based baseline and our PhyScene3D in the dining table scene.}
    \label{tab:diffusion}
    \resizebox{\columnwidth}{!}{%
    \begin{tabular}{ccccc}
    \toprule
    Model & QPR ($\tau=7$) & GPT Score & SCR & ACR \\
    \hline
    DiffuScene & 0 & 5.95 & 100 & 15.6 \\
    PhyScene3D (Ours) & 0.242 & 8.58 & 73.62 & 5.96 \\
    \bottomrule
    \end{tabular}%
    }
\end{table}

To investigate the robustness of CTRC, we analyze performance across varying difficulty levels stratified by object density (Figure~\ref{fig:difficulty_curve}). As scene complexity increases from Level 1 (sparse) to Level 5 (dense), the performance of the Reference data drops sharply, dropping to 0.0\% QPR at Level 4, indicating that human annotators struggle to maintain collision-free consistencies in dense environments. While GPT-4o and MesaTask exhibit sharp performance degradation at higher difficulties (dropping to 19.0\% and 7.1\% at Level 5, respectively), PhyScene3D demonstrates superior resilience, maintaining a QPR of 40\% at Level 5. This suggests that the hierarchical decomposition in CTRC allows the model to manage combinatorial complexity effectively, preventing the ``cascading errors" observed in non-hierarchical baselines.

Beyond regression baselines, our method also demonstrates superior performance compared to the diffusion-based baseline. As shown in Table~\ref{tab:diffusion}, DiffuScene~\cite{tang2024diffuscene}. completely fails under dense tabletop constraints, yielding 0\% QPR, 100\% scene-wise collision, and 5.95 GPT Score on the representative Dining Table subset. While diffusion models excel at global distribution modeling, their lack of explicit 3D physical boundary enforcement renders them incapable of handling dense local interactions. This validates that tabletop scene generation requires a physics-aligned framework rather than pure generative modeling.

Beyond standard benchmarks, a robust generative model must exhibit zero-shot generalization to novel semantic contexts. To validate this, we evaluate PhyScene3D on four unseen scenes: \textit{Cashier Counter, Nightstand, Side Table, and TV Stand}. From Table~\ref{tab:ood_result}, the \textit{MesaTask} suffers a severe performance drop (QPR dropping to 1.01\%), revealing that end-to-end regression models tend to overfit the absolute spatial distributions of the training set. In contrast, PhyScene3D demonstrates remarkable adaptability. The introduction of \textbf{CTRC} yields a $10\times$ improvement in QPR (10.6\%) over the baseline, suggesting that our hierarchical relative representation captures \textit{topological invariants} (e.g., ``placement on a surface" is structurally identical whether the surface is a desk or a nightstand). Furthermore, the integration of \textbf{PADA} boosts QPR to 29.1\%. This confirms that physical laws are universal. By grounding the model in fundamental constraints (gravity, non-penetration) rather than dataset statistics, PADA enables the agent to construct valid layouts even in environments it has never seen.

\subsubsection{Qualitative Analysis}
\label{subsec:qualitative}

\begin{figure*}[h]
    \centering
    \includegraphics[width=1\linewidth]{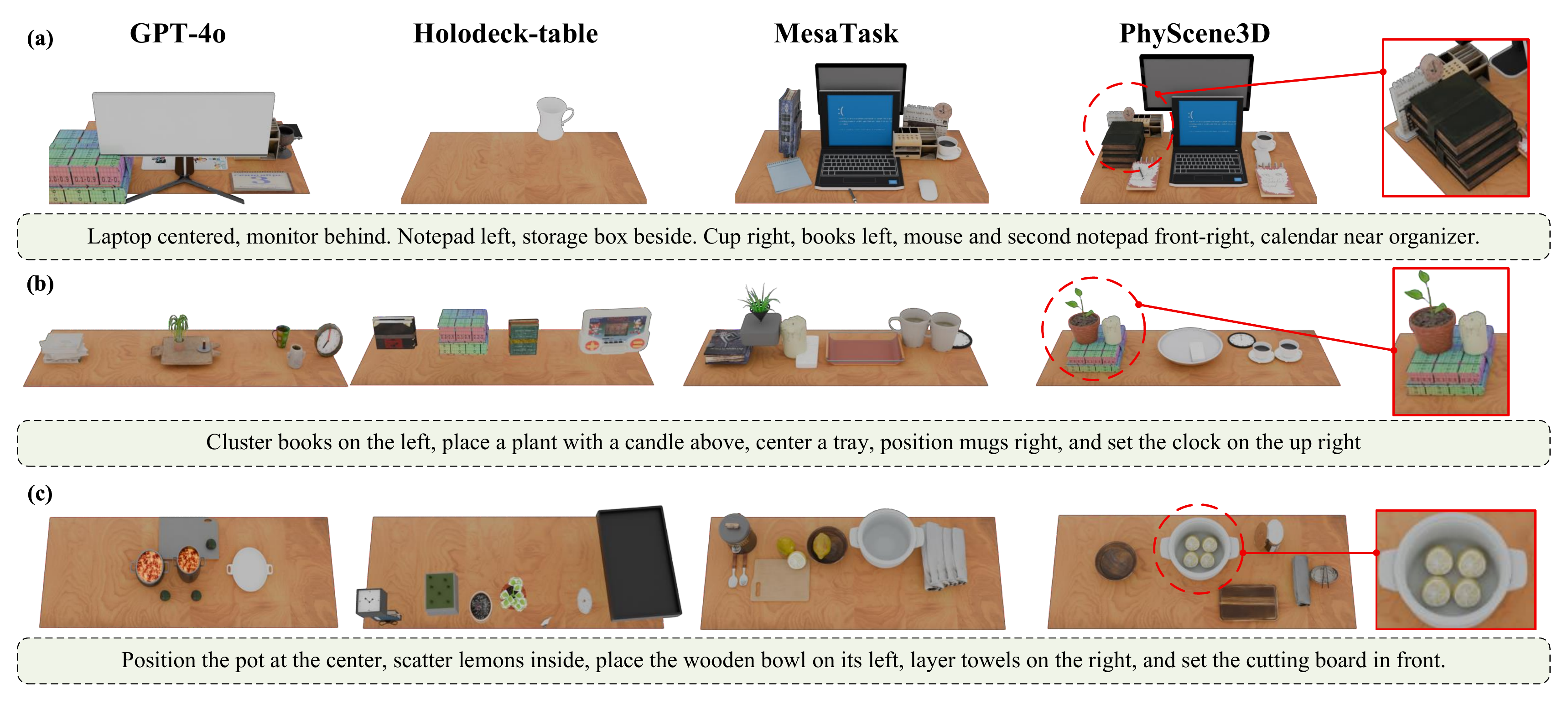}
    \caption{Qualitative comparison of generated desktop scenes. \textbf{GPT-4o} often hallucinates floating objects or intersects volumes. \textbf{Holodeck} produces sparse, unnatural arrangements due to symbolic bottlenecks. \textbf{MesaTask} mimics training noise with visible collisions. \textbf{PhyScene3D (Ours)} generates hierarchically dense, collision-free layouts with precise container-content relationships.}
    \label{fig:vis}
\end{figure*}

We visually compare the generated scenes in Figure~\ref{fig:vis} to better interpret the observed numerical gains. The examples illustrate distinct failure modes in baseline methods, which are effectively circumvented by PhyScene3D.

For instance, while GPT-4o captures high-level semantics, it struggles with fine-grained spatial awareness. Objects frequently float mid-air or exhibit severe inter-penetration (e.g., closely packed small objects interpenetrate, see in Figure~\ref{fig:vis} (a)). This highlights the limitations of even powerful foundational models in precise 3D spatial reasoning when lacking an explicit grounding mechanism like PADA.

In contrast, Holodeck generates physically valid but semantically limited scenes. The layouts are often sparse and rigid, failing to naturally cluster objects. This reflects the inherent trade-off in compressing rich linguistic instructions into symbolic code: while the solver satisfies physical constraints, it misses subtler nuances, such as the ``cluttered" yet realistic arrangement of objects on a typical tabletop.

Similarly, MesaTask produces scenes with visually plausible density but inherits the physics errors present in its training set. We observe unstable stacks (e.g., the book, plant, and candle balanced precariously, see in Figure~\ref{fig:vis} (b)) and intersecting meshes, which align with our quantitative findings. Standard regression methods, as demonstrated, tend to blindly replicate noise from the Reference distribution, resulting in these artifacts.

In comparison, our method demonstrates the unique benefits of the CTRC-PADA combination. In complex stacking scenarios (e.g., ``lemon in the pot" or ``a plant with a candle above the books", see in Figure~\ref{fig:vis}), PhyScene3D accurately resolves dependency chains: containers are placed first, appropriately sized, and their contents are grounded within them. The AABB-based placement scheme ensures coherence in local assemblies, while TTO fine-tunes contact points to high precision. The resulting scenes achieve a tidily dense layout that closely mimics human organization, one that is structurally sound and semantically consistent.

\begin{table}
    \centering
    \caption{Robotic grasping success rates in the ManiSkill simulator.}
    \label{tab:robot_sr}
    \resizebox{0.9\columnwidth}{!}{%
    \begin{tabular}{ccc}
    \toprule     
    Robotic Task Success Rate & IID scene & OOD scene\\
    \hline
    MesaTask & 4.6\% & 3.5\% \\
    PhyScene3D (Ours) & 50.4\% & 14.1\% \\
    \bottomrule
    \end{tabular}%
    }
\end{table}

Beyond static layout quality, the ultimate test of an interactive scene is whether it supports physical manipulation. We import generated layouts into the ManiSkill simulator and train robotic agents for cluttered target retrieval. Agents trained on MesaTask baseline scenes, which contain severe interpenetration, suffer from simulation instabilities during initialization and achieve only 4.6\% success in IID settings and 3.5\% in OOD. In contrast, agents trained on PhyScene3D layouts succeed in 50.4\% of IID episodes and 14.1\% OOD. This dramatic improvement directly validates the qualitative observations in Figure~\ref{tab:robot_sr}: the collision-free stacks and precise container content relationships produced by our method are not merely visually convincing, but provide a stable physical foundation that enables reliable robotic interaction.

\begin{table*}[t]
    \centering
    \setlength{\tabcolsep}{4pt}
        \caption{Ablation study on the MesaTask-CTRC benchmark. 
        \textbf{CTRC} denotes SFT on hierarchical sequences; \textbf{+GRPO} denotes SFT followed by pure reinforcement learning without anchors; \textbf{+TTO} denotes post-hoc optimization applied to SFT outputs; \textbf{+PADA*} denotes SFT with RL using raw SFT outputs as semantic anchors; and \textbf{+PADA} denotes SFT with RL using TTO-refined layouts as anchors.
        }
        
    \begin{tabular}{lccccccccccc}
    \toprule
    \multirow{2}{*}{Model}  & \multicolumn{3}{c}{Quality Pass Rate} & \multicolumn{6}{c}{GPT Score} & \multicolumn{2}{c}{Collision Ratio} \\
    \cmidrule(lr){2-4}\cmidrule(lr){5-10}\cmidrule(lr){11-12}
    & $\tau=7$ & $\tau=8$ & $\tau=9$ & CwT & OSR & PPI & LCR & OV & Avg. & Scene-wise$\downarrow$ & Asset-wise$\downarrow$\\
    \midrule
    \rowcolor{gray!10}
    Qwen-3 8B & 18.5\% & 15.8\% & 12.7\% & 7.74 & 9.44 & 9.83 & 8.21 & 9.11 & 8.86 & 81.1\% & 8.50\% \\
    \rowcolor{gray!10}
    + CTRC  & 20.3\% & 19.0\% & 14.9\% & 7.71 & 9.43 & 9.83 & 8.19 & 9.09 & 8.85 & 79.3\% & 7.46\% \\
    \midrule
    Qwen-3 VL 8B & 19.2\% & 17.2\% & 13.7\% & 7.76 & 9.28 & 9.86 & 8.18 & 9.12 & 8.84 & 80.1\% & 8.18\%\\
    + CTRC  & 21.6\% & 20.0\% & 15.8\% & \textbf{7.91} & \textbf{9.51} & \underline{9.88} & \textbf{8.34} & \textbf{9.18} & \textbf{8.96} & 77.8\% &7.60\% \\
    + GRPO & 28.8\% & 22.8\% & 14.1\% & 7.11 & 9.23 & 9.73 & 7.67 & 8.60 & 8.47 & 68.9\% & 6.82\% \\
    + TTO & \underline{38.0}\% & \underline{34.8}\% & \underline{24.2}\% & 7.71 & 9.36 & 9.79 & 8.17 & 9.12 & 8.83 & \underline{60.5}\% & \underline{4.86}\%\\
    \rowcolor{red!10}
    + PADA* & 34.2\% & 29.9\% & 19.8\% & 7.51 & 9.38 & 9.79 & 8.00 & 8.98 & 8.73 & 64.5\% & 5.63\% \\  
    \rowcolor{red!10}
    + PADA & \textbf{46.5}\% & \textbf{39.1}\% & \textbf{27.3}\% & \underline{7.85} & \underline{9.45} & \textbf{9.91} & \underline{8.29} & \underline{9.15} & \underline{8.93} & \textbf{41.6}\% & \textbf{3.86}\%\\
    \bottomrule
    \end{tabular}
    \label{tab:ablation}
\end{table*}

\subsection{Ablation Study}
To disentangle the contributions of each component in PhyScene3D, we conduct a component-wise analysis using the Qwen-3 VL 8B backbone, as shown in Table~\ref{tab:ablation}. 


The transition from a text-only LLM (Qwen-3 8B) to a VLM (Qwen-3 VL 8B) yields modest improvements (+0.7\% QPR at $\tau=7$), confirming that visual pre-training aligns latent representations closer to 3D spatial concepts. However, the introduction of the CTRC generates a more significant leap (+2.4\% QPR), validating that explicitly modeling scene hierarchy and relative coordinates provides a robust inductive bias. This structural constraint reduces geometric hallucinations (Asset Collision drops from 8.18\% to 7.60\%) even without explicit physical feedback, proving that hierarchical planning is fundamental to handling dense layouts.

We observe a critical trade-off when introducing physical supervision. While naive reinforcement learning (\textbf{+GRPO}) significantly improves physical plausibility (Asset Collision decreases to 6.82\%), it suffers from severe reward hacking, evidenced by a sharp drop in Average GPT Score (8.96 to 8.47). This drift manifests not only in lower GPT scores but also in reduced spatial diversity, as shown in the Appendix~\ref{app:diversity}. The model learns to minimize collisions by generating sparse, simplistic layouts that violate semantic instructions. Test-time optimization (\textbf{+TTO}) effectively mitigates this by correcting artifacts post-hoc, achieving a strong QPR of 38.0\%. Its value lies in generating offline training targets for PADA, not in serving as a practical deployment-time solution. 
However, TTO is computationally expensive during deployment and limited by initialization quality; it serves as local repair rather than fundamental correction of the generative policy.

Our proposed \textbf{PADA} framework shows the benefit of internalizing these constraints during training. By using TTO-refined layouts as semantic anchors, PADA surpasses the inference-only TTO baseline (\red{46.5}\% vs 38.0\% QPR) and achieves the lowest collision rate (\red{3.86}\%). By distilling the TTO engine into the generative policy rather than applying local post-hoc repair, PADA achieves a better trade-off between semantic compactness and physical validity that neither pure GRPO nor isolated TTO can achieve. This indicates that the VLM has successfully ``learned physics" from the TTO teacher, generating initial proposals that are better globally than what local optimization can achieve from noisy initializations. 

Comparing anchor strategies, using the raw SFT output as an anchor (\textbf{PADA*}) restores semantic fidelity (GPT Score 8.73 \& 5.63\% collision), which means semantic anchors steer the model’s exploration beyond GRPO (GPT Score 8.47 \& 6.82\% collision), ensuring both semantic coherence and physical plausibility. Yet it still falls short of performance from the TTO-refined anchor (3.86\% collision). Furthermore, tabletop scenes are extremely dense, typically containing 10 to 20 objects with complex contact relationships. Requiring absolute zero collision across every scene remains extremely difficult for end-to-end generative models; however, reducing local asset-wise interpenetration to 3.86\% represents a substantial breakthrough in physical fidelity. Thus, the TTO-refined anchor (\textbf{PADA}) offers a better Pareto frontier, combining the semantic richness of the SFT model with the rigorous physical validity of the SDF engine.

\section{Conclusion and Future Work}
In this work, we presented \textbf{PhyScene3D}, a framework that reformulates 3D tabletop scene generation from a static regression problem into a \textit{Human-Mimetic Constructive Process}. By combining the \textbf{CTRC} for hierarchical planning with \textbf{(PADA)} for geometric grounding, our method overcomes the limitations of noisy supervision, generating environments that are physically more plausible than the human-annotated data on which it is trained. We argue that the integration of high-fidelity \textbf{differentiable physics engines} and robust \textbf{semantic reward models} will be critical to bridging the sim-to-real gap. These advances aim to enable robotic agents to perceive, reason, and manipulate within the complex and unstructured physical world.


\clearpage

\section*{Acknowledgement}
This work is supported by National Key Research and Development Program of China (2024YFE0203100), in part by the National Natural Science Foundation of China under Grant No. 62325605, No. 62572498, No. 62536010 and No. 62322608, in part by the open research fund of Pengcheng Laboratory under Grant 2025KF1B0050, in part by the Guangdong Basic and Applied Basic Research Foundation under Grant No. 2025A1515011874. This work was supported in part by the Key Development Project of the Artificial Intelligence Institute, Sun Yat-sen University, under Grant 2025RGZN009. We also thank the National Supercomputer Center in Guangzhou for computational support.

\section*{Impact Statement}
This paper presents work whose goal is to advance the field of Machine Learning. There are many potential societal consequences of our work, none of which we feel must be specifically highlighted here.

\bibliography{reference}
\bibliographystyle{icml2026}

\newpage
\appendix
\onecolumn
\section{Dataset}
\label{app:dataset}

We curate \textbf{MesaTask-CTRC}, a dataset explicitly designed to train hierarchical spatial reasoning, derived from the MesaTask-10k repository~\cite{haomesatask}. While the original dataset provides high-quality 3D assets and raw layout JSONs, it lacks the semantic annotations required for instruction-following tasks. Our pipeline consists of three stages:

\subsection{Geometric Relationship Extraction}
To construct the ground-truth Hierarchical Scene Graph (HSG), we parse the raw absolute coordinates using geometric heuristics. Let $A$ and $B$ be two objects with bounding boxes $\mathcal{B}_A, \mathcal{B}_B$. We define relationships as follows:
\begin{itemize}
    \item \textbf{Containment ($\mathtt{in}$):} $A$ is inside $B$ if the volume ratio $Vol(B)/Vol(A) \ge 1.5$ and the intersection over union on the XY-plane $IoU_{xy} \ge 0.9$.
    \item \textbf{Support ($\mathtt{on}$):} $A$ is on $B$ if the vertical gap $|z_{min}(A) - z_{max}(B)| < \delta$ (where $\delta=1\text{cm}$) and $IoU_{xy} > 0.3$.
    \item \textbf{Proximity ($\mathtt{near}$):} If no vertical relationship exists, we classify spatial relations (left/right/front/back) based on the vector connecting their centroids projected onto the table plane.
\end{itemize}

\subsection{VLM-Augmented Instruction Synthesis}
We employ a ``Caption-then-Instruct" pipeline using the current state-of-the-art Vision-Language Model (VLM) to bridge the modality gap:
\begin{enumerate}
    \item \textbf{Multi-View Captioning:} We render the scene from four orthographic angles (Front, Top, Left, Right). The VLM receives these images along with the textual Scene Graph and generates a dense description $\mathcal{D}$ ($<200$ words) focusing on spatial configurations (e.g., \textit{"A cluttered study desk where a laptop sits centrally on a stand, flanked by a stack of books to the right..."}).
    \item \textbf{Instruction Generation:} Conditioned on $\mathcal{D}$, we prompt the VLM to act as a user and issue a goal-oriented command $\mathcal{I}$ ($<30$ words). We enforce an "actionable" constraint: the instruction must imply the arrangement rather than listing coordinates (e.g., \textit{"Set up a workstation for coding with reference materials within reach"}).
\end{enumerate}
This process yields 9,429 training pairs and 866 benchmark pairs. The benchmark is manually verified to ensure no data leakage.

\subsection{Benchmark Statistics}
Our benchmark is rigorously stratified to evaluate model performance across varying degrees of complexity. Table~\ref{tab:benchmark_stats} details the distribution of the 866 test samples across 6 functional scene categories and 5 difficulty levels based on the number of objects ($N$). Level 5 ($N \ge 17$) represents extreme clutter scenarios that challenge the stability of physical reasoning.

Additionally, we built the OOD benchmark, including 4 desktop types, e.g., Cashier Counter, Nightstand, Side Table, and TV Stand. This OOD benchmark counts 50 cases per desktop type, totally 200 cases.

\begin{table}[h]
    \centering
    \caption{Statistics of the MesaTask-CTRC Benchmark. Samples are categorized by scene function and difficulty level (object count range).}
    \label{tab:benchmark_stats}
    \vspace{0.2cm}
    \resizebox{\linewidth}{!}{
    \begin{tabular}{lcccccc}
    \toprule
    \textbf{Scene Type} & \textbf{Level 1 (1-4)} & \textbf{Level 2 (5-8)} & \textbf{Level 3 (9-12)} & \textbf{Level 4 (13-16)} & \textbf{Level 5 (17+)} & \textbf{Total} \\
    \midrule
    Coffee Table    & 6 & 47 & 41 & 15 & 6 & 115 \\
    Office Table    & 3 & 54 & 57 & 25 & 10 & 149 \\
    Dressing Table  & 7 & 59 & 60 & 22 & 8 & 156 \\
    Dining Table    & 7 & 63 & 56 & 26 & 11 & 163 \\
    Kitchen Counter & 5 & 56 & 52 & 16 & 5 & 134 \\
    Bathroom Vanity & 7 & 59 & 58 & 23 & 2 & 149 \\
    \midrule
    \textbf{Total}  & \textbf{35} & \textbf{338} & \textbf{324} & \textbf{127} & \textbf{42} & \textbf{866} \\
    \bottomrule
    \end{tabular}
    }
\end{table}

\subsection{Integrated Data Generation Algorithm}
The construction of the Hierarchical Scene Graph (HSG), the Unified Relative Coordinate System, and the Object-Level Sub-instructions is a simultaneous, unified process. We employ a recursive parsing strategy centered on anchor identification. Algorithm~\ref{alg:data_gen} outlines this procedure.

\begin{algorithm}[h]
   \caption{Integrated Generation of HSG, Relative Coordinates, and Instructions}
   \label{alg:data_gen}
\begin{algorithmic}[1]
   \STATE {\bfseries Input:} Raw Layout $\mathcal{L} = \{ (o_i, \mathbf{b}^{abs}_i) \}$
   \STATE {\bfseries Output:} Sequence of instructions $\mathcal{I}_{seq}$ and relative targets $\mathcal{T}_{seq}$
   \STATE {\bfseries Function} \textsc{ProcessNode}($parent, candidates$):
   \STATE \quad $\mathcal{I}_{seq} \gets [], \mathcal{T}_{seq} \gets []$
   \STATE \quad Sort $candidates$ by $z$-axis (bottom-up) and proximity to $parent$
   \FOR{$child \in candidates$}
       \STATE \quad \textcolor{gray}{// 1. Determine Geometric Relationship}
       \STATE \quad $relation \gets \text{ClassifyRelation}(child, parent)$ \COMMENT{in / on / near}
       
       \STATE \quad \textcolor{gray}{// 2. Compute Relative Coordinate (Unified System)}
       \STATE \quad $\mathbf{b}^{rel} \gets \text{ComputeRelativeBBox}(child.\mathbf{b}^{abs}, parent.\mathbf{b}^{abs}, relation)$
       
       \STATE \quad \textcolor{gray}{// 3. Generate Sub-Instruction}
       \IF{$parent$ is Root}
           \STATE $text \gets \text{Template}(\text{"Place } child.name \text{ at } \mathbf{b}^{rel}.\text{coarse\_loc on the table."})$
       \ELSE
           \STATE $text \gets \text{Template}(\text{"Place } child.name \text{ } relation \text{ } parent.name \text{."})$
       \ENDIF
       
       \STATE \quad Append $(text)$ to $\mathcal{I}_{seq}$, Append $(\mathbf{b}^{rel})$ to $\mathcal{T}_{seq}$
       
       \STATE \quad \textcolor{gray}{// 4. Recursive Expansion (Stacking)}
       \STATE \quad $grandchildren \gets \text{FindChildren}(child, \mathcal{L})$
       \STATE \quad $\mathcal{I}_{sub}, \mathcal{T}_{sub} \gets \textsc{ProcessNode}(child, grandchildren)$
       \STATE \quad Extend $\mathcal{I}_{seq}$ with $\mathcal{I}_{sub}$, $\mathcal{T}_{seq}$ with $\mathcal{T}_{sub}$
   \ENDFOR
   \STATE \quad \textbf{return} $\mathcal{I}_{seq}, \mathcal{T}_{seq}$
   
   \STATE {\bfseries Main Process:}
   \STATE $Anchor \gets \text{FindClosestToCenter}(\mathcal{L})$
   \STATE Run \textsc{ProcessNode}(RootTable, $[Anchor]$)
   \STATE Run \textsc{ProcessNode}(RootTable, $RemainingObjects$)
\end{algorithmic}
\end{algorithm}

\textbf{Reference Example.} Below is a sample output segment illustrating the mapping from instruction to relative geometric targets:

\begin{mdframed}[backgroundcolor=verbgray,hidealllines=true,innerleftmargin=3pt,innerrightmargin=3pt]
\begin{verbatim}
The tabletop features a woven basket positioned in the back right corner, facing 
forward. To the left of the basket, there are two lotion bottles placed closely 
together in the back left area. The first lotion bottle, which is a soapenser lo-
tion bottle, is situated behind the second lotion bottle, which is also facing f-
orward. The arrangement creates a densely packed area on the left side of the ta-
ble, while the right side remains relatively empty, dominated by the basket. The 
overall layout is asymmetric, with the basket providing a focal point on the rig-
ht and the lotion bottles creating a cluster on the left, leaving ample space in 
the center of the tabletop.

**Object List**: woven basket (1), lotion bottle (1), soapenser lotion bottle (1)
**Scene Graph:**
# ========== Stack-level Horizontal Relationships ==========
(1_soapenser_lotion_bottle_0, left of, 10_lotion_bottle_1)
# ========== Within-Stack Relationships ==========

# ========== Object Absolute Properties ==========
(2_woven_basket_0, face to, front)
\n(2_woven_basket_0, is at, behind_right)
(1_soapenser_lotion_bottle_0, face to, front)
(1_soapenser_lotion_bottle_0, is at, behind_left)
(10_lotion_bottle_1, face to, front)
(10_lotion_bottle_1, is at, behind_left)

**Sub-instruction:**
[
    {
        "step": 1, 
        "instance_id": "10_lotion_bottle_1", 
        "instruction": "Place 10_lotion_bottle_1 at the left back of center of t
                        he table, approximately 28cm from center.", 
        "reference_objects": []
    }, 
    {
        "step": 2, 
        "instance_id": "1_soapenser_lotion_bottle_0", 
        "instruction": "Place 1_soapenser_lotion_bottle_0 to the left of 10_loti
                        on_bottle_1.", 
        "reference_objects": ["10_lotion_bottle_1"]
    }, 
    {
        "step": 3, 
        "instance_id": "2_woven_basket_0", 
        "instruction": "Place 2_woven_basket_0 near 10_lotion_bottle_1, about 53
                        cm away.", 
        "reference_objects": ["10_lotion_bottle_1"]
    }
]

**Scene Layout:**
{
    "objects": [
                   {
                       "layer": 0, 
                       "instance_id": "root_zone", 
                       "type": "zone", 
                       "bbox": [0.0, 80.5, 0.0, 41.9, 0.0, 0.0], 
                       "z_rotation": 0.0
                    }, 
                    {
                       "layer": 1, 
                       "instance_id": "2_woven_basket_0", 
                       "parent_id": "root_zone", 
                       "relation_type": "on_surface", 
                       "description": "Rectangular woven basket of natural fiber
                                       s, light brown color with darker accents 
                                       and distinctive texture.", 
                       "relative_bbox": [57.05, 78.75, 20.7, 35.5, 0.0, 15.0], 
                       "relative_rotation_deg": 0.0
                    }, 
    
                    {
                       "layer": 1, 
                       "instance_id": "1_soapenser_lotion_bottle_0", 
                       "parent_id": "root_zone", 
                       "relation_type": "on_surface", 
                       "description": "Tall, cylindrical, smooth surface, pastel 
                                       colors, slightly tapered base, labeled wi
                                       th brand information.", 
                       "relative_bbox": [2.75, 12.85, 30.9, 36.1, 0.0, 22.0], 
                       "relative_rotation_deg": 0.0
                    }, 
                    {
                       "layer": 1, 
                       "instance_id": "10_lotion_bottle_1", 
                       "parent_id": "root_zone", 
                       "relation_type": "on_surface", 
                       "description": "Tall, cylindrical, smooth surface, pastel 
                                       colors, slightly tapered base, labeled wi
                                       th brand information.", 
                       "relative_bbox": [13.8, 21.6, 31.75, 35.65, 0.0, 16.2], 
                       "relative_rotation_deg": 0.0}], 
                       "item_placement_zone": [0.0, 80.5, 0.0, 41.9]
                    }
\end{verbatim}
\end{mdframed}

\section{Implementation Details.}
\label{app:implementation}

\subsection{Model Architecture and Training}
We utilize \textbf{Qwen3-VL-8B}~\cite{bai2025qwen3vltechnicalreport} as our backbone. All experiments are conducted on 640 GB VRAM within a cluster of 8$\times$ GPUs using DeepSpeed ZeRO-2 optimization.

\textbf{Stage 1: Supervised Fine-Tuning (SFT).}
We perform full-parameter fine-tuning to align the model with the CTRC format.
\begin{itemize}
    \item \textbf{Optimizer:} AdamW with $\beta_1=0.9, \beta_2=0.95$.
    \item \textbf{Learning Rate:} $1 \times 10^{-5}$ with Cosine Decay (warmup ratio 0.03).
    \item \textbf{Batch Size:} Global batch size of 64.
    \item \textbf{Epochs:} 5.
\end{itemize}

\textbf{Stage 2: Physically-Aware RL (PADA).}
We freeze the backbone and train LoRA adapters to inject physical gradients.
\begin{itemize}
    \item \textbf{LoRA Configuration:} Rank $r=16$, Alpha $\alpha=32$, applied to \texttt{q\_proj, k\_proj, v\_proj, o\_proj, gate\_proj, up\_proj, down\_proj}.
    \item \textbf{Algorithm:} Group Relative Policy Optimization (GRPO).
    \item \textbf{Group Size ($G$):} 4 samples per prompt.
    \item \textbf{Learning Rate:} $5 \times 10^{-4}$ (constant).
    \item \textbf{Batch Size:} Global batch size of 128.
    \item \textbf{Epochs:} 1.
    \item \textbf{KL Coefficient ($\beta$):} 0.04 (initial) with adaptive adjustment.
    \item \textbf{Semantic Anchor Weight ($\gamma$):} 0.1 (weight of the semantic consistency loss).
\end{itemize}

Note that in \textbf{PADA*}, we apply a mask when computing the CE loss for anchor samples, supervising only the semantic part. This is because the BBox and rotation values need to be explored by GRPO and should not be aligned with pseudo-labels that contain collisions.

\textbf{Reward Formulation.} 
To guide the policy towards physically plausible and structurally valid generation, we design a discontinuous reward function $R(\mathcal{L})$ that strictly penalizes formatting errors and physical violations while incentivizing collision-free solutions. Let $\mathcal{E}_{pen}$ denote the total penetration depth (accumulated collision energy) calculated by our SDF engine. The reward is defined as:

\begin{equation}
    R(\mathcal{L}) = 
    \begin{cases} 
    -100 & \text{if Parsing Failed (Format Error)} \\
    -\min(0.01 \cdot \mathcal{E}_{pen}, 50) & \text{if } \mathcal{E}_{pen} > 0 \\
    +50 & \text{if } \mathcal{E}_{pen} = 0
    \end{cases}
\end{equation}

where $\mathcal{E}_{pen} = \sum_{i<j} \text{pen}(i,j)$ sums the penetration depths of all colliding pairs. This formulation ensures that: 
1) \textbf{Format Compliance} is the prerequisite ($-100$ penalty); 
2) \textbf{Collision Minimization} provides a dense gradient for partial failures, clipped at $-50$ to prevent reward explosion; 
3) \textbf{Perfect Physics} yields a sparse positive signal ($+50$) to reinforce the generated "Golden Samples."

\subsection{SDF and TTO Configuration}
The Physics Engine preprocesses assets into $64^3$ SDF grids. During Test-Time Optimization (TTO) and Anchor Generation:
\begin{itemize}
    \item \textbf{Optimization Steps:} 100 iterations using Adam ($lr=0.01$).
    \item \textbf{Loss Weights:} $\lambda_{sdf}=10.0$ (Collision), $\lambda_{rel}=5.0$ (Structure), $\lambda_{reg}=1.0$ (Drift).
    \item \textbf{Collision Threshold:} A penetration depth $>1\text{cm}$ is considered a collision.
\end{itemize}

\section{Metric}

\label{app:metrics}

\subsection{Physical Plausibility}
We quantify physical violations using the precomputed SDFs. Let $\mathcal{S}$ be a generated scene with $N$ objects.
\begin{itemize}
    \item \textbf{Asset-wise Collision Rate (ACR):} The percentage of object pairs $(i, j)$ that intersect:
    $$ ACR = \frac{1}{|\mathcal{D}|} \sum_{s \in \mathcal{D}} \frac{2}{N(N-1)} \sum_{i<j} \mathbb{I}(\text{pen}(i,j) > 0\text{cm}) $$
    \item \textbf{Scene-wise Collision Rate (SCR):} The percentage of scenes containing at least one collision:
    $$ SCR = \frac{1}{|\mathcal{D}|} \sum_{s \in \mathcal{D}} \mathbb{I}(\exists i,j : \text{pen}(i,j) > 0\text{cm}) $$
\end{itemize}

The collision metric relies on the function $\text{pen}(i, j)$. We define $\text{pen}(i, j)$ as the \textbf{Penetration Depth}, representing the maximum distance one object's geometry overlaps into another's. Formally, for two signed distance fields $\phi_i$ and $\phi_j$, the penetration depth is approximated by the maximum negative SDF value of points from object $i$ queried in object $j$'s field. We set a tolerance threshold of $0\text{cm}$ to account for mesh surface noise; only penetrations exceeding this value contribute to the Scene Collision Rate (SCR) and Asset Collision Rate (ACR).

\subsection{Semantic Fidelity (GPT-Score)}
We employ GPT-4o as an external evaluator. The model is provided with the text instruction and the generated JSON layout (parsed into natural language descriptions of relative positions). It scores the layout on a scale of 1-10 across five dimensions:
\begin{enumerate}
    \item \textbf{Consistency with Task (CwT):} Does the scene serve the function implied by the instruction?
    \item \textbf{Object Size Reasonableness (OSR):} Are objects scaled realistically relative to each other?
    \item \textbf{Placement Plausibility \& Intersections (PPI):} Are objects placed logically (e.g., mouse near keyboard) without obvious overlapping descriptions?
    \item \textbf{Layout Coherence \& Realism (LCR):} Does the global arrangement reflect a human-like workspace?
    \item \textbf{Object Visibility (OV):} Are smaller objects obscured or hidden inside unrelated containers?
\end{enumerate}
The \textbf{Average GPT Score} is the arithmetic mean of these five components.

\subsection{Quality Pass Rate (QPR)}
To measure the practical usability of the generated data, we define QPR as the intersection of semantic success and physical validity. For a threshold $\tau$ (e.g., 7):
$$ QPR(\tau) = \frac{1}{|\mathcal{D}|} \sum_{s \in \mathcal{D}} \left( \mathbb{I}(\text{Score}_{avg}(s) \ge \tau) \times \mathbb{I}(\text{collisions}(s) = 0) \right) $$
This metric is strict; a scene must be both semantically high-quality and physically perfect to pass.

\section{More Result}

\begin{figure}[h!]
    \centering
    \includegraphics[width=0.5\linewidth]{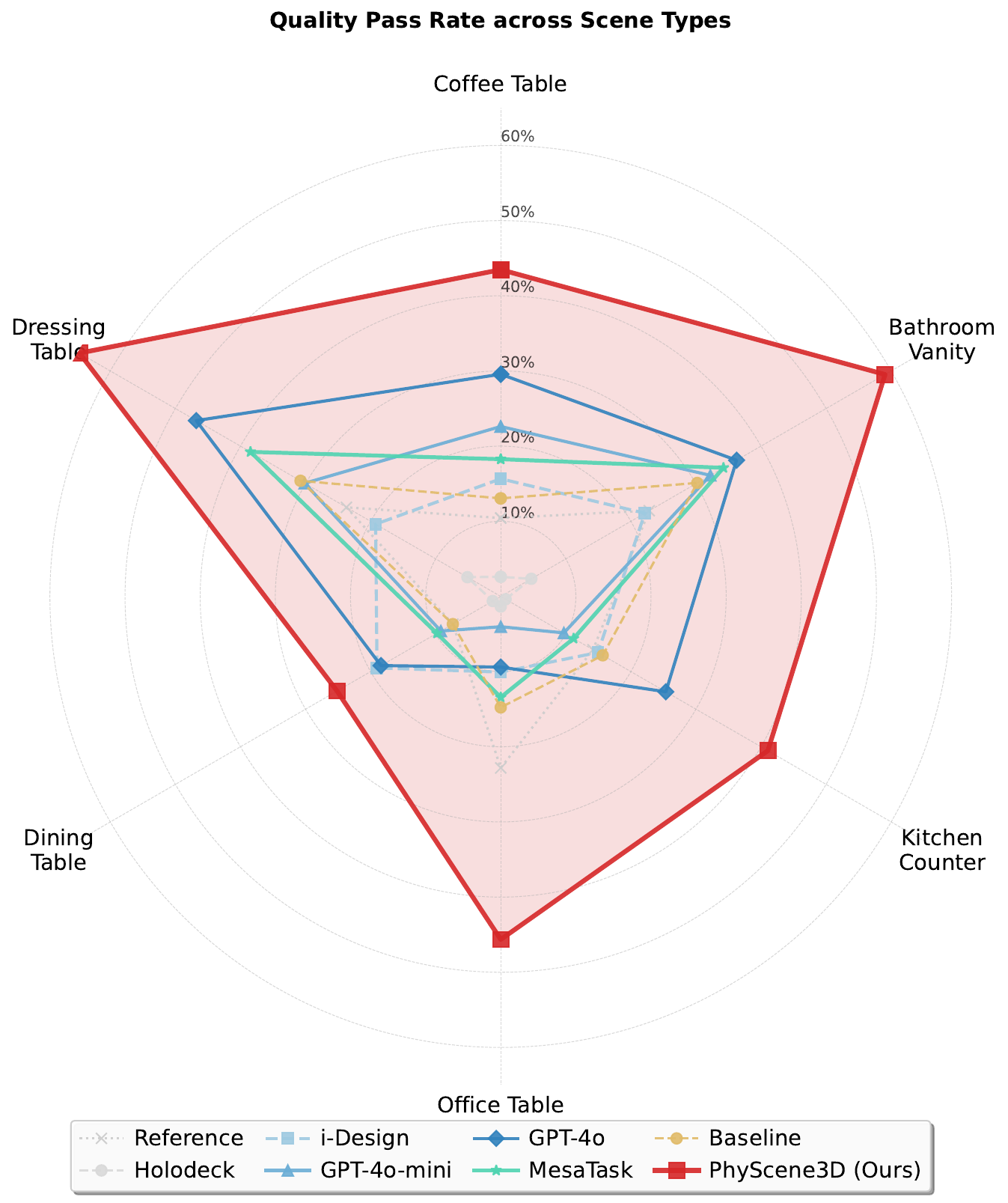}
    \caption{QPR ($\tau=7$) performance of each method across different scene settings.}
    \label{fig:radar_chart}
\end{figure}
Additionally, we further assess generalization across six distinct functional scenarios (Figure~\ref{fig:radar_chart}). PhyScene3D consistently encloses the performance polygons of competing methods. Notably, in the Office Table scenario, characterized by high object density and strict logical constraints (e.g., monitors aligned, peripherals anchored), our method achieves a QPR of more than \red{45}\%, significantly outperforming MesaTask (13.4\%) and Reference (22.8\%). Even in the Dressing Table scenario, which requires precise clutter arrangement, PhyScene3D achieves \red{65}\%, effectively doubling the performance of GPT-4o-Mini (30.1\%). This cross-scenario robustness validates that our method learns a generalized physical-semantic grammar rather than overfitting to specific layout patterns.

\section{Prompt}

\begin{mdframed}[
  backgroundcolor=verbgray,
  hidealllines=true,
  innerleftmargin=3pt,
  innerrightmargin=3pt,
  frametitle={\sffamily\bfseries Prompt 1. Description Generation},
  frametitleaboveskip=2pt,
  frametitlebelowskip=2pt
  ]
\begin{verbatim}
The given images are the front view (in which the table is symmetric) and the sc-
ene graph. Only describe the layout on the tabletop; do not include the descript-
ion of the table. Please describe the scene in detail, including the objects (do 
not describe the color/material/color of the object), their positions, and the o-
verall layout of the desktop, including empty and densely packed are as Output t-
he description in a paragraph between the token <description>, no more than 200 
words.

## Scene Graph
{scene_graph}
\end{verbatim}
\end{mdframed}

\begin{mdframed}[
  backgroundcolor=verbgray,
  hidealllines=true,
  innerleftmargin=3pt,
  innerrightmargin=3pt,
  frametitle={\sffamily\bfseries Prompt 2. Instruction Generation},
  frametitleaboveskip=2pt,
  frametitlebelowskip=2pt
  ]
\begin{verbatim}
<description>{description}<description>

If I provide the assets and tabletops needed to place the task, how can I guide 
you through a short and concise instruction (less than 30 words) to arrange this 
layout (the layout that matches the above scenario description)?

You said that the proposed instruction should not refer to the description, such 
as "as described", "Arrange items as described" and so on. Because in the real 
world, I only provide commands, assets, and desktops.

Output the instruction between the token <instruction>
\end{verbatim}
\end{mdframed}

\begin{mdframed}[
  backgroundcolor=verbgray,
  hidealllines=true,
  innerleftmargin=3pt,
  innerrightmargin=3pt,
  frametitle={\sffamily\bfseries Prompt 3. Layout Generation by Zero-shot LLM},
  frametitleaboveskip=2pt,
  frametitlebelowskip=2pt
  ]
\begin{verbatim}
You are tasked with generating a scene graph layout for a tabletop arrangement 
based on the following instructions and parameters. 

### TASK:
Your primary goal is to output a **VALID JSON OBJECT** that strictly follows the 
specified format between the `<json>` tokens. This JSON must represent the scene 
graph layout for objects placed on a tabletop.

### INPUT PARAMETERS:
1. **Tabletop Information:**
   - The tabletop's dimensions are defined by the "item_placement_zone" with the 
   following boundaries:  
     **item_placement_zone**: <item_placement_zone>
     Units are in centimeters (cm).

2. **Placement Instructions:**
   - Follow the natural language instructions to determine the placement of obje-
     cts:  
     **Instructions**: "<instructions>"

3. **Objects Cluster:**
   - The objects to be placed on the tabletop are listed below. Each object must 
     have:  
     - A **unique instance name**.
     - **Reasonable dimensions** (width, depth, height in cm).
     - A **position** within the specified tabletop boundaries.
     - A **z_rotation** (rotation around the vertical axis in radians). If not s-
     pecified, assume `z_rotation = 0.0`.
     - A **short description** that explains the object's purpose or appearance.  
     **Objects cluster**: <Objects cluster>

### OUTPUT REQUIREMENTS:
- You must output a **VALID JSON OBJECT** strictly between the `<json>` tokens.  
  (not <json>xxx</json>)
- The JSON must have the following structure:

<json>
{
    "item_placement_zone": [
        x_min,
        x_max,
        y_min,
        y_max
    ],
    "objects": [
        {
            "instance": "unique_object_name",
            "z_rotation": float,
            "size": [width, depth, height],
            "position": [x, y, z],
            "description": "short description of the object"
        },
        ...
    ]
}
<json>

### DETAILED GUIDELINES:
1. **Object Placement:**
   - Ensure that all objects are placed within the boundaries of the `item_place-
     ment_zone`.
   - Follow the placement logic specified in the instructions. For example:  
     - If an object should be placed in a corner, position it near the respective 
       boundary.  
     - If objects need to be grouped closely, ensure their positions are adjacent 
       within reasonable spacing (e.g., 2-5 cm).  

2. **Object Properties:**
   - Assign **realistic dimensions** to each object based on its type. For insta-
     nce:  
     - A woven basket might have dimensions like `[20.0, 15.0, 12.0]` (width, de-
       pth, height).  
     - A lotion bottle might have dimensions like `[5.0, 5.0, 15.0]`.
   - Use **intuitive and consistent instance names** (e.g., "woven_basket_0", "l-
     otion_bottle_1").
   - Set the `z_rotation` to `0.0` unless the object is explicitly rotated in the 
     instructions.

3. **Avoid Overlapping:**
   - Ensure objects do not overlap or extend beyond the tabletop boundaries.
   - Maintain logical spacing between objects unless the instructions specify ot-
     herwise.

4. **Avoid Suspension:**
    - Ensure that the items are firmly placed on the table. The position of each 
      asset is the center coordinate, so the z value cannot be zero. If it is on 
      the table, it should be half of the asset's height.
    - If there is a stacking situation, the z coordinate of the upper asset sho-
      uld be the sum of the z coordinate of the lower asset and half of the hei-
      ght of both assets.

5. **Output Validation:**
   - Ensure the JSON is valid by checking for correct syntax (e.g., no trailing 
     commas, proper nesting of keys and values).

IMPORTANT:
The JSON output MUST be placed between the <json> tokens.
If you encounter ambiguity, use logical reasoning to infer the most likely pla-
cement or dimensions.
DO NOT include any additional text or explanations outside the <json> tokens in 
the output.
\end{verbatim}
\end{mdframed}

\begin{mdframed}[
  backgroundcolor=verbgray,
  hidealllines=true,
  innerleftmargin=3pt,
  innerrightmargin=3pt,
  frametitle={\sffamily\bfseries Prompt 4. GPT Score (System prompt)},
  frametitleaboveskip=2pt,
  frametitlebelowskip=2pt
  ]
\begin{verbatim}
You are an expert evaluator for 3D desktop scene layouts. Your task is to analyze 
desktop scenes and provide a detailed assessment based on specific criteria. Ple-
ase carefully examine the provided front and perspective views of the desktop sc-
ene, along with the task description. 
Analyze how well the scene layout aligns with the intended task. Consider both t-
he visible objects and their arrangement in relation to the task requirements. 
You must provide numerical scores (1-10) for each criterion along with brief exp-
lanations to justify your ratings. 
Be objective and consistent in your evaluation across different scenes.
\end{verbatim}
\end{mdframed}

\begin{mdframed}[
  backgroundcolor=verbgray,
  hidealllines=true,
  innerleftmargin=3pt,
  innerrightmargin=3pt,
  frametitle={\sffamily\bfseries Prompt 5. GPT Score (user prompt)},
  frametitleaboveskip=2pt,
  frametitlebelowskip=2pt
  ]
\begin{verbatim}
You are an expert at evaluating desktop scene layouts. Given both a front and a 
perspective view of a desktop scene, and a description of the tasks that can be 
performed on the desktop, please rate the scene`s quality on a scale from 1 (poo-
r) to 10 (excellent) according to the following criteria. For each criterion, co-
nsider both views:  

1. **Consistency with Task:** Does the scene layout (objects present, their arra-
ngement and relevance) align well with the provided task description (environment
, objects, goals)? 
- High score (7-10): All key objects are present, their arrangement is entirely 
logical and directly contributes to the task. The scene perfectly reflects the t-
ask requirements. 
- Mid score (4-6): Most key objects are present and generally align with the task
, but there might be minor inconsistencies or some less relevant elements. 
- Low score (1-3): Significant deviations from the task description. Important o-
bjects are missing, or the arrangement is largely unrelated or illogical for the 
task.  

2. **Object Size Reasonableness:** Are the sizes of the objects in the scene hig-
hly realistic, both relative to each other and to the overall desktop environmen-
t? Are they consistent across all objects? 
- High score (7-10): All objects have highly realistic and perfectly proportiona-
te sizes. No inconsistencies are noticeable. 
- Mid score (4-6): Most objects have reasonable sizes, but there might be slight 
or occasional inaccuracies in proportion for some items. 
- Low score (1-3): Multiple or glaring inconsistencies in object sizes. Some obj-
ects are obviously and significantly too large or too small, severely impacting 
realism.  

3. **Placement Plausibility & Intersections:** Are objects placed stably and nat-
urally on surfaces (e.g., not floating)? Are there any signs of unnatural inters-
ection or penetration between objects? Objects should appear physically grounded 
and interact realistically. 
- High score (7-10): All objects rest naturally and stably on surfaces. No aphys-
ical interactions or intersections are visible. Objects are well-supported. 
- Mid score (4-6): Most objects are placed plausibly, but there might be minor, 
subtle issues like slight floating or minimal, non-critical intersections. 
- Low score (1-3): Obvious or frequent issues with object placement. Objects flo-
at, unnaturally overlap, or clearly penetrate each other, indicating a lack of p-
hysical realism.  

4. **Layout Coherence & Realism:** Does the overall arrangement look highly func-
tional, convincingly realistic, and typical for the task context? Does it avoid 
being overly staged, unnaturally sparse, or chaotically cluttered? 
- High score (7-10): Layout is highly functional, convincingly realistic, and we-
ll-suited for the described task. The scene feels authentic and natural. 
- Mid score (4-6): The layout is generally coherent and functional, but might la-
ck some fine-tuning for optimal realism or could appear somewhat staged. 
- Low score (1-3): Layout is chaotic, illogical, too empty, overly cluttered, or 
looks clearly artificial and unrealistic for the task.

5. **Object Visibility:** Are important objects mentioned in the task easily and 
unambiguously identifiable in at least one of the views? Are they sufficiently w-
ell-lit and resolved? 
- High score (7-10): All key objects are clearly and unambiguously visible and i-
dentifiable. Their details are well-resolved. 
- Mid score (4-6): Most important objects are visible, but some might require cl-
oser inspection to identify. 
- Low score (1-3): Critical objects are very difficult to identify, or completel-
y missing from view, hindering task understanding.  

**Task Description:**  

{task_description}

**Please provide a single score (1-10) for each criterion.** 
**Strictly output in the following JSON format with no additional text:**  

{  
    "Evaluation": [ 
        {
            "criterion": "Consistency with Task", 
            "explanation": "Your detailed explanation here", 
            "score": X
        }, 
        {  
            "criterion": "Object Size Reasonableness", 
            "explanation": "Your detailed explanation here", 
            "score": X
        }, 
        {  
            "criterion": "Placement Plausibility & Intersections", 
            "explanation": "Your detailed explanation here", 
            "score": X
        }, 
        {  
            "criterion": "Layout Coherence & Realism", 
            "explanation": "Your detailed explanation here", 
            "score": X 
        }, 
        {  
            "criterion": "Object Visibility", 
            "explanation": "Your detailed explanation here", 
            "score": X 
        }
    ] 
}  

**Where X is an integer score from 1 to 10. Do not output anything else.**
\end{verbatim}
\end{mdframed}

\section{Generative Diversity Analysis}
\label{app:diversity}

Conventional reinforcement learning is known to cause mode collapse, and conservative layouts with reduced positional variance are frequently produced. To verify whether PADA preserves generative diversity, the number of distinct asset types and the spatial standard deviations along the X and Y axes were measured for both IID and OOD scenarios. As shown in Tables~\ref{tab:diversity_iid} and~\ref{tab:diversity_ood}, a noticeable decline in diversity is observed for pure GRPO without semantic anchors, reflected by smaller positional standard deviations. In contrast, spatial diversity comparable to that of the SFT baseline and MesaTask is maintained by PADA, which employs semantic anchors refined by TTO together with KL divergence constraints.

\begin{table}[h]
\centering
\caption{Generative diversity comparison on IID scenes.}
\label{tab:diversity_iid}
\begin{tabular}{lcc}
\toprule
Model & Asset Types & Pos Std X / Y (cm) \\
\midrule
GPT-4o & 1207 & 36.24 / 14.6 \\
MesaTask & 844 & 37.17 / 11.94 \\
SFT Only & 932 & 35.11 / 11.4 \\
Pure GRPO (No Anchor) & 924 & 33.83 / 11.31 \\
PADA (Ours) & 927 & 36.99 / 12.03 \\
\bottomrule
\end{tabular}
\end{table}

\begin{table}[h]
\centering
\caption{Generative diversity comparison on OOD scenes.}
\label{tab:diversity_ood}
\begin{tabular}{lcc}
\toprule
Model & Asset Types & Pos Std X / Y (cm) \\
\midrule
GPT-4o & 303 & 29.99 / 15.64 \\
MesaTask & 287 & 33.63 / 11.45 \\
SFT Only & 290 & 32.71 / 10.8 \\
Pure GRPO (No Anchor) & 281 & 31.47 / 10.14 \\
PADA (Ours) & 287 & 33.7 / 12.61 \\
\bottomrule
\end{tabular}
\end{table}


\begin{table}[!h]
\centering
\caption{Latency and VRAM overhead per stage.}
\label{tab:latency}
\resizebox{\textwidth}{!}{
\begin{tabular}{lcccc}
\toprule
Stage / Module & Time & VRAM per GPU (Peak) & VRAM per GPU (Avg) & Remarks \\
\midrule
SFT Training & 31.8 min/epoch & 34.5 GB & 32.3 GB & Full parameter fine tuning \\
TTO (Offline Data Preparation) & 1.64 s/scene & 8.4 GB & 5.4 GB & Executed only once; does not occupy training time \\
SDF Reward Computation (RL Stage) & 305.6 ms/batch & 35.5 GB & 33.4 GB & Highly parallelized operator; minimal additional overhead \\
PADA Overall Training (GRPO) & 5.3 hours/batch & 80.7 GB & 60.3 GB & LoRA adapters only \\
Inference (Generation) & 0.67 s/scene & 71.7 GB & 62.4 GB & Single forward pass; extremely fast \\
\bottomrule
\end{tabular}
}
\end{table}

\section{Computational Overhead and Latency Analysis}
\label{app:latency}

The computational overhead introduced by the physics engine is incurred solely during training. At inference time, only a single forward pass is required by the model to produce a complete layout, and the need for iterative LLM calls or lengthy denoising processes is thereby eliminated. The latency and memory consumption for each stage, measured on a cluster of eight GPUs, are summarized in Table~\ref{tab:latency}.

\section{Robustness to Collision Thresholds}
\label{app:threshold}

The 81.5\% collision rate in the reference data was strictly computed using high precision mesh SDFs rather than coarse AABB approximations. To examine robustness under varying evaluation criteria, the scene wise collision rate was evaluated again with penetration thresholds of 0.0 cm, 0.5 cm, and 1.0 cm.

As shown in Table~\ref{tab:threshold}, a substantially lower collision rate is still achieved by PhyScene3D when the threshold is relaxed to 1.0 cm, compared with both MesaTask and the reference data. This indicates that the improvement stems from genuine physical geometric reasoning rather than fitting to bounding box errors.

\begin{table}[!h]
\centering
\caption{SCR under varying collision thresholds.}
\label{tab:threshold}
\begin{tabular}{lcccc}
\toprule
Collision Threshold & Ref & GPT-4o & MesaTask & Ours \\
\midrule
0.0 cm (Strict) & 81.5\% & 68.9\% & 78.3\% & 41.6\% \\
0.5 cm & 76.1\% & 61.9\% & 76.1\% & 38.4\% \\
1.0 cm & 63.4\% & 52.3\% & 61.8\% & 34.1\% \\
\bottomrule
\end{tabular}
\end{table}

\section{Failure Cases and Limitations of AABB}
\label{app:limitation}

The limitations of AABB representations are fully acknowledged, particularly for objects with extreme irregular shapes. For instance, occupied volumes that are significantly larger than the actual physical extents are assigned to objects with large cavities or C shaped structures, such as woven baskets with tall handles or over ear headphones. In rare cases, overly conservative behavior is induced in the PADA engine during optimization, and surrounding objects are pushed farther apart than necessary, producing locally sparse layouts. 


\end{document}